%% file: acl_latex.tex
\documentclass[11pt]{article}

\usepackage[preprint]{acl}

\usepackage{subcaption}
\usepackage{multirow}
\usepackage{times}
\usepackage{latexsym}
\usepackage{lipsum}
\usepackage[T1]{fontenc}

\usepackage[utf8]{inputenc}

\usepackage{microtype}

\usepackage{inconsolata}

\usepackage{graphicx}

\usepackage{listings}

\usepackage{hyperref}
\usepackage{url}
\usepackage{booktabs}
\usepackage{graphicx}
 \usepackage{float}
\usepackage{pifont}
\usepackage{newunicodechar}

\newunicodechar{Δ}{\ensuremath{\Delta}}
\newunicodechar{★}{\ding{72}}

\usepackage{booktabs}

\usepackage{caption}    
\usepackage{float} 
\usepackage[capitalize, nameinlink]{cleveref} 

\usepackage{booktabs}
\usepackage{siunitx}
\usepackage{adjustbox} 

\usepackage{xcolor}

\usepackage[table]{xcolor}
\usepackage{hhline}

\usepackage{graphicx}

\definecolor{aurocCol}{RGB}{235,245,255}   
\definecolor{accCol}{RGB}{245,235,255}     

\definecolor{scblue}{RGB}{220,240,255}
\definecolor{vcgreen}{RGB}{220,255,220}
\definecolor{vcscpink}{RGB}{255,220,230}

\lstdefinestyle{prompt}{
  basicstyle=\ttfamily\scriptsize,
  breaklines=true,
  frame=single,
  framerule=0.4pt,
  rulecolor=\color{gray},
  backgroundcolor=\color{gray!3},
  tabsize=1,
  captionpos=b,   
}

\newcommand{\tpm}{\mathbin{\scalebox{0.8}{$\pm$}}}

\title{How Uncertainty Estimation Scales with Sampling in Reasoning Models}

\author{
Maksym Del, Markus Kängsepp, Marharyta Domnich \\
\bf{Ardi Tampuu}, Lisa Yankovskaya, Meelis Kull, Mark Fishel \\
Institute of Computer Science\\ University of Tartu
}

\begin{document}
\maketitle

\begin{abstract}
Uncertainty estimation is critical for deploying reasoning language models, yet remains poorly understood under extended chain-of-thought reasoning. We study parallel sampling as a fully black-box approach using verbalized confidence and self-consistency. Across three reasoning models and 17 tasks spanning mathematics, STEM, and humanities, we characterize how these signals scale.

Both self-consistency and verbalized confidence scale in reasoning models, but self-consistency exhibits lower initial discrimination and lags behind verbalized confidence under moderate sampling. Most uncertainty gains, however, arise from signal combination: with just two samples, a hybrid estimator improves AUROC by up to $+12$ on average and already outperforms either signal alone even when scaled to much larger budgets, after which returns diminish. These effects are domain-dependent: in mathematics, the native domain of RLVR-style post-training, reasoning models achieve higher uncertainty quality and exhibit both stronger complementarity and faster scaling than in STEM or humanities.
\end{abstract}

\section{Introduction}

Uncertainty estimation is a central component of reliable machine learning. It enables selective prediction, risk-aware decision-making, and safe integration of models into larger systems. Reasoning language models (RLMs) extend standard large language models by extending test-time computation through extended chain-of-thought deliberation. These models have demonstrated strong performance and are increasingly deployed in numerous domains, including those where reliable uncertainty estimation is particularly important. Although extended reasoning is primarily used to improve answer accuracy, it also reveals uncertainty information that can be extracted in a fully black-box manner, e.g., by eliciting verbalized confidence from a single model run \cite{lin2022teaching}.

Beyond elicitation, uncertainty estimation can also exploit parallel sampling, which queries the same prompt multiple times to obtain multiple samples. Variation across samples can be used to estimate uncertainty, for example, through consistency in final responses \cite{wang2022self} or by aggregating introspective signals across samples \cite{xiong_can_2024}. Moreover, it also becomes possible to estimate uncertainty with a combined signal, where introspection and consistency are added together \cite{chen-mueller-2024-quantifying}. 

Parallel sampling may unlock higher-quality uncertainty estimates, but in RLMs each additional sample is particularly costly because it entails another full chain-of-thought trace. We therefore need to understand how uncertainty estimation scales with the number of samples, how quickly improvements saturate, and whether combining signals yields further gains. Although uncertainty estimation in reasoning language models has recently received attention, existing studies evaluate only the single-sample setting \cite{zeng-etal-2025-thinking, yoon2025reasoning, mei2025reasoninguncertaintyreasoningmodels}, leaving multi-sample behavior and interactions between introspective and agreement-based signals uncharacterized.

\paragraph{Why transfer of insights from standard non-reasoning language models is not guaranteed.}
In contrast, uncertainty estimation for standard large language models has been studied in multi-sample settings. Prior work examines agreement-based metrics such as self-consistency, as well as combinations of introspective and agreement-based signals \cite{xiong_can_2024, chen-mueller-2024-quantifying}, exploring to what extent these approaches can be effective in standard non-reasoning models. However, a direct transfer of these findings cannot be naively assumed due to fundamental differences between reasoning and non-reasoning models that alter how uncertainty is produced:
\begin{itemize}
    \item First, a single sample in a reasoning model involves extended intra-sample deliberation, during which a larger number of candidate solutions may be internally explored and discarded. Consequently, a single sample may already capture some uncertainty that would only emerge through additional samples in standard models, making the marginal value of further sampling unclear.
    
    \item Second, as introspection is performed on traces potentially surfacing a larger number of hypotheses, it may change how introspective and agreement-based signals interact. Signals that were complementary under shallow reasoning may exhibit different synergies or become more redundant once uncertainty is partially resolved during deliberation.
    
    \item Third, the way uncertainty signals in reasoning language models scale and complement each other can vary systematically across domains compared to standard language models. Reasoning models undergo a key post-training procedure via reinforcement learning with verifiable rewards (RLVR), but this is primarily limited to mathematical domains.
\end{itemize}

\paragraph{Research goal and scope.}
\emph{Our goal is to understand how black-box uncertainty estimation behaves under parallel sampling in reasoning language models. We study how introspection-based and agreement-based uncertainty signals scale with the number of test-time samples, how quickly their gains saturate, and how these signals interact under the extended chain-of-thought reasoning across domains.}

\paragraph{Contributions and Findings}

\begin{itemize}
    \item \textbf{Contribution: The first systematic benchmark of uncertainty scaling in reasoning language models.}
    We present the first large-scale empirical study of how black-box uncertainty estimation scales with parallel sampling in reasoning language models. Across three models and 17 tasks spanning mathematics, STEM, and humanities, we evaluate verbalized confidence (VC), self-consistency (SC), and their combination under a unified bootstrap-based protocol, establishing a comprehensive empirical reference for uncertainty estimation under extended chain-of-thought reasoning (Section~\ref{sec:results}).
    
    \item \textbf{Finding I: Self-consistency and verbalized confidence both scale in reasoning language models, but self-consistency lags behind.}
    Under parallel sampling, VC provides a strong baseline in reasoning language models and can be further improved via scaling, gaining up to $+10.1$ AUROC in mathematics from $K{=}1$ to $K{=}8$, but saturating earlier and yielding smaller gains ($+4$--$5$ AUROC) in STEM and humanities. In contrast, self-consistency (SC) starts substantially below VC at comparable budgets (e.g., $70.5$ vs.\ $73.5$ AUROC at $K{=}2$ in mathematics) and, while it improves with additional samples, does not overtake VC within the tested range.

    \item \textbf{Finding II: Signal complementarity dominates scaling of either signal alone.}
    Combining introspection- and agreement-based uncertainty signals yields substantially larger gains than increasing the sampling budget for VC or SC individually. With just two samples, the hybrid estimator (SCVC) improves AUROC by $+12.9$ points in mathematics and $+6.4$ points in STEM and humanities relative to single-sample VC, and already outperforms VC or SC scaled to $K{=}8$ across all domains.

    \item \textbf{Finding III: Uncertainty signals scale fastest and combine most effectively in mathematics.}
    Scaling behavior is strongly domain-dependent: mathematics exhibits both larger immediate gains from combining signals and more sustained improvement with additional samples (SCVC: $+2.7$ AUROC from $K{=}2{\rightarrow}5$ and $+1.5$ from $K{=}5{\rightarrow}8$), whereas STEM and humanities saturate earlier (approximately $+1$--$2$ AUROC beyond $K{=}2$). This aligned behavior across VC, SC, and their combination is consistent with reasoning language models being most optimized for mathematical reasoning.

\end{itemize}

\section{Preliminaries}

\subsection{Methods}

We study black-box uncertainty estimation from parallel sampling. For each input, we sample $K$ independent model samples, where each sample corresponds to a full model execution producing a reasoning trace and a final answer $\{(r_i, a_i)\}_{i=1}^K$. A sample can additionally yield a verbalized confidence score $c_i \in [0,1]$ if so prompted. Let $\hat{a}$ denote the majority-vote answer among $\{a_i\}_{i=1}^K$; ties are broken uniformly at random.

\paragraph{Verbalized confidence (VC).}
Verbalized confidence uses the model's explicit confidence outputs as a black-box uncertainty signal \cite{kadavath_language_2022,lin_teaching_2022}. We prompt the model to report a numeric confidence value alongside its final answer using a standardized epistemic elicitation prompt. Specifically, we adopt an Epistemic Elicitation (EpEL; \citealp{tian_just_2023}) instruction that encourages the model to reflect on its subjective certainty in the provided answer without revisiting the solution. The same EpEL prompt is used consistently across all domains, including mathematics, STEM, and humanities. We provide complete prompts in Appendix~\ref{app:prompts}, and analyze the impact of alternative confidence elicitation variants in Section~\ref{sec:vc_variants}.

We rescale the reported confidence value to $[0,1]$ by dividing by $100$.
With $K$ sampled traces, we average confidence over those predicting the majority answer $\hat{a}$:
\[
\mathrm{VC_{avg}} \;=\; \frac{1}{|\{i : a_i=\hat{a}\}|}\sum_{i : a_i=\hat{a}} c_i.
\]
We also refer to $VC_{avg}$ simply as $VC$ when it is clear that it is being sampled.

\paragraph{Self-consistency (SC).}
Self-consistency \citep{wang2022self} estimates confidence from agreement across $K$ sampled samples:
\[
\mathrm{SC} \;=\; \frac{1}{K}\sum_{i=1}^K \mathbf{1}[a_i=\hat{a}].
\]

\paragraph{Combined signal (SCVC).}
SC and VC can also be combined \cite{chen-mueller-2024-quantifying, xiong_can_2024, huang2024calibratinglongformgenerationslarge, rivera2024combining}. In this study we explore the following minimal combination:
\[
\mathrm{SCVC} \;=\; \lambda \cdot \mathrm{SC} + (1-\lambda)\cdot \mathrm{VC}_{\mathrm{avg}},
\]
with $\lambda=0.5$ by default; we vary $\lambda$ in Section~\ref{sec:analysis_ablations}.

\subsection{Tasks and Models}
\label{sec:tasks_models}

\paragraph{Tasks.}
We evaluate uncertainty estimation in three task families: mathematics, STEM (excluding math), and humanities, covering 17 tasks in total (Table~\ref{tab:task_families}). Mathematical tasks are a key in-domain setting for RLVR-style post-training of reasoning models \citep{ma2025generalreasoneradvancingllmreasoning}. Our math suite includes \texttt{GSM8K} \citep{cobbe2021trainingverifierssolvemath}, \texttt{AIME 2024 \& 2025} (30 problems each, combined into a single task) \citep{aime_problems_solutions}, and the Math subset of \texttt{MMLU-Pro} \citep{wang2024mmlupro}.

For non-mathematical reasoning, we include \texttt{GPQA Diamond} \citep{rein2024gpqa} and multiple subject areas from \texttt{MMLU-Pro} \citep{wang2024mmlupro}, spanning STEM (Health, Biology, Chemistry, Economics, Physics, Computer Science, Engineering) and humanities (Psychology, Law, Business, History, Philosophy, Other). Tasks are mainly multiple-choice; math tasks additionally include free-form numeric answers, such as GSM8K and AIME. The number of examples per task is reported in Table~\ref{tab:task_families}.

\paragraph{Models.}
We use three open-source reasoning models: \texttt{gpt-oss-20b}\footnote{\url{https://huggingface.co/openai/gpt-oss-20b}} (with reasoning effort set to high), \texttt{Qwen3-30B-A3B}\footnote{\url{https://huggingface.co/Qwen/Qwen3-30B-A3B}}, and \texttt{DeepSeek-R1-8B}\footnote{\url{https://huggingface.co/deepseek-ai/DeepSeek-R1-0528-Qwen3-8B}}. \texttt{gpt-oss-20b} and \texttt{Qwen3-30B-A3B} are mixture-of-experts models trained with Reinforcement Learning with Verifiable Rewards (RLVR). \texttt{DeepSeek-R1-8B} is a dense model obtained by fine-tuning an \texttt{8B Qwen base} on reasoning traces from DeepSeek-R1 \citep{deepseekr1}. We select these mid-sized models to retain strong reasoning performance while enabling robust parallel sampling with up to 100 samples per question. All models support context windows of at least 131K tokens, enabling extended chain-of-thought reasoning.

\paragraph{Generation configuration.}
We allow models to generate up to 60K tokens per sample, which fits within the 131K context window and leaves room for an additional confidence estimation pass of up to 60K tokens (needed for Section~\ref{sec:vc_variants}). All evaluations are performed using the vLLM framework\footnote{\url{https://github.com/vllm-project/vllm}}. We use generation hyperparameters recommended by the model authors: temperature = 1.0 and top-$p$ = 1.0 for \texttt{gpt-oss-20b}, and temperature = 0.6 and top-$p$ = 0.95 for \texttt{Qwen3-30B-A3B} and \texttt{DeepSeek-R1-8B}.

\begin{table}[]
\centering
\input{tables_readonly/task_families_examples}
\caption{Task families and constituent tasks used in our evaluation, along with the number of examples per task.}
\label{tab:task_families}
\end{table}

\subsection{AUROC for confidence evaluation}
We evaluate confidence signals by how discriminative they are about correctness. A signal is \emph{discriminative} if correct answers tend to receive higher scores than incorrect ones. A signal is \emph{calibrated} if predictions with assigned confidence $p$ are correct about $p$ of the time. Calibration metrics such as ECE \citep{guo2017calibration} or the Brier score \citep{brier1950verification} require meaningful probabilistic interpretation of the numeric scale. They depend on scale alignment, which is undesirable here because 1) VC is self-reported and often mis-scaled \cite{tian_just_2023, xiong_can_2024, chen_reasoning_nodate}, 2) SC is an agreement statistic with coarse, $K$-dependent precision. Because both signals may be monotonic but mis-calibrated, discrimination provides
a scale-invariant evaluation of uncertainty quality, we therefore choose AUROC \citep{hanley1982auc}. Formally, AUROC measures the probability that a randomly chosen correct example
receives a higher confidence score than a randomly chosen incorrect example.  

In our setting, the binary labels are answer correctness and the scores are the confidence signal values; AUROC can be interpreted as the probability that, for a randomly chosen correct and incorrect example, the correct one receives a higher confidence score, with 0.5 corresponding to random ranking and higher values indicating better discrimination.

\subsection{Bootstrap evaluation protocol}

Our goal is to estimate AUROC at a target sampling budget $K$ under stochastic decoding where the variance from temperature sampling can be large. Specifically, for each question in our datasets, we first generate a pool of $R$ independent samples (reasoning chain, answer, and confidence) using repeated decoding. We use $R=10$ for all main results aiming to cover moderate-sampling regime which is most practically relevant for reasoning models where each chain of though sample is long and expensive.. 
Individual generations that do not follow the required answer or confidence format are discarded, and questions with no valid samples are removed (less than 1\% of examples).

For each question $q$, we first generate a pool of $R$ independent samples 
$\{(r_{q,i}, a_{q,i}, c_{q,i})\}_{i=1}^{R}$ under stochastic decoding.
To estimate performance at sampling budget $K$ while accounting for decoding variance,
we perform $B$ bootstrap draws as follows. In each draw, we uniformly sample $K$ elements per question without replacement,
compute VC, SC, or SCVC from these samples, and evaluate AUROC over all questions.
Task-level AUROC is macro-averaged within domain, and model-level averages are computed within the same draw.
We report the mean across $B$ draws with 95\% percentile intervals. The per-question pool serves as a Monte Carlo approximation to the model's decoding distribution, so repeatedly drawing $K$ elements yields plausible datasets that one would obtain by re-decoding the model, reducing variance on estimates while making maximal use of the generated samples. Repeating this procedure over $B$ bootstrap draws yields a distribution of aggregated AUROC values.

\begin{table*}[t]
\centering

\input{tables_readonly/auroc_task_tables_combined_summary_tasks}

\caption{AUROC of verbalized confidence (VC), self-consistency (SC), and their combination (SCVC) across task families at varying sampling budgets $K$. Tasks are primarily from MMLU-Pro (Math, STEM, and Social Sciences/Humanities); the STEM domain also includes GPQA Diamond, which is not part of MMLU-Pro. Rows report per-task AUROC macro-averaged over models, and ``Average (tasks)'' denotes the mean $\pm$ bootstrap standard deviation macro-averaged over models and tasks within task family (domain).}

\label{tab:auroc_task_tables_combined_summary}
\end{table*}

\begin{figure*}[t]
    \centering
    \includegraphics[width=\linewidth]{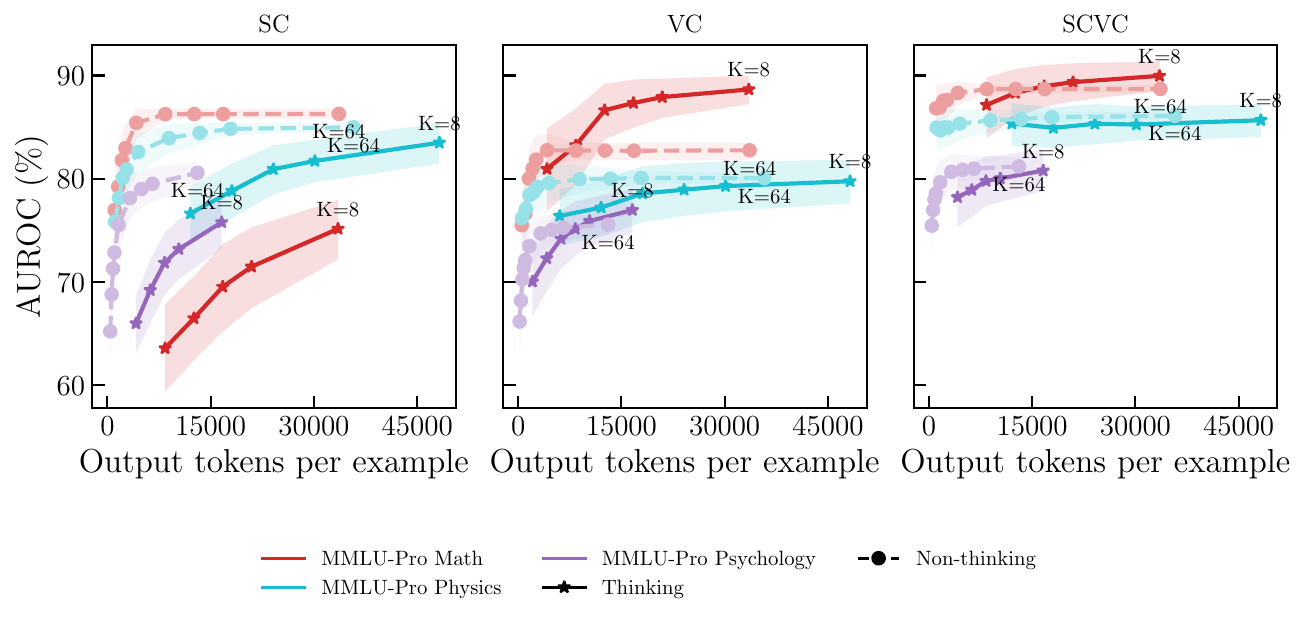}
    \caption{%
     Direct comparison (AUROC vs.\ cost across datasets) between extended thinking (gpt-oss-20b-high) and shallow thinking (gpt-oss-20b-low).
    }
    \label{fig:gpt20b_cost_accuracy_grid}
\end{figure*}

\section{Experimental Results}
\label{sec:results}

\subsection{Finding I: SC and VC both scale in RLMs, but SC demonstrates lower sample efficiency in the low-budget regime}
\label{sec:results:observations}

\paragraph{Verbalized confidence at one sample.}
Verbalized confidence provides strong uncertainty discrimination in reasoning language models, 
even with a single sample, consistent with recent findings on single-sample introspection in RLMs \cite{zeng-etal-2025-thinking, yoon2025reasoning, mei2025reasoninguncertaintyreasoningmodels} and illustrated in Figure~\ref{fig:gpt20b_cost_accuracy_grid}. At $K{=}1$, VC achieves high AUROC on most tasks, including 78.8 on MMLU-Pro Math, 68.7 on GSM8K, 73.8 on average across STEM tasks, and 68.5 across humanities. The lower average in mathematics is driven primarily by AIME (66.4 AUROC), a small and highly sensitive benchmark, whereas larger math tasks exhibit substantially stronger VC. We analyze VC prompt sensitivity across tasks in Section~\ref{sec:vc_variants} and show that simplified elicitation substantially improves VC on AIME without affecting downstream conclusions.

\paragraph{Scaling verbalized confidence.}
Beyond the single-sample regime, VC can be scaled via parallel sampling by aggregating confidence across samples associated with the majority answer, a setting that has not been systematically characterized for reasoning language models. VC scales substantially in mathematics, improving from 71.3 at $K{=}1$ to 81.4 at $K{=}8$ (+10.1 AUROC), but exhibits much smaller gains in STEM and humanities (+4.6 and +4.1, respectively). Across non-mathematical domains, VC scaling is strongly front-loaded and largely saturates by $K{\approx}5$.

\paragraph{Self-consistency.}
Self-consistency (SC) behaves differently in reasoning language models than in standard language models. At comparable sampling budgets, SC starts substantially below VC across all domains: at $K{=}2$, SC attains 70.5 AUROC in mathematics, 66.6 in STEM, and 63.3 in humanities, consistently trailing VC at the same budget. The apparent competitiveness of SC on AIME reflects the unusually low VC on that task; beyond AIME, the gap is pronounced, with SC@2 reaching only 65.9 on MMLU-Pro Math and 63.3 on GSM8K, compared to 78.8 and 68.7 for VC@1 on the same benchmarks. A controlled comparison using GPT-OSS-20B under matched token budgets further shows that SC is markedly weaker under extended reasoning than under shallow (LLM-like) generation across math, physics, and psychology, while VC remains strong and scales reliably (Figure~\ref{fig:gpt20b_cost_accuracy_grid}). Although SC improves steadily with additional samples, it does not overtake VC within the tested range up to $K{=}8$.

\paragraph{Summary.}
Taken together, these results provide the first systematic characterization of individual uncertainty signals under parallel sampling in reasoning language models. Verbalized confidence is a strong baseline that can be further improved via sampling, particularly in mathematics, while self-consistency emerges more slowly and remains weaker at comparable budgets. These findings establish the individual behavior of introspection- and agreement-based signals in RLMs, independently of their combination.

\subsection{Finding II: Signal complementarity dominates scaling of either signal alone.}
\label{sec:results:complementarity}

\paragraph{Claim.}
In reasoning language models, combining introspection- and agreement-based uncertainty signals provides substantially larger gains than increasing the sampling budget for either signal alone. While verbalized confidence and self-consistency each benefit from additional samples, their combination unlocks most of the attainable uncertainty quality at the smallest sampling budget where both signals become available.

\paragraph{Evidence.}
Table~\ref{tab:auroc_task_tables_combined_summary} shows that combining verbalized confidence and self-consistency yields large, strongly front-loaded improvements across all domains. In mathematics, SCVC at $K{=}2$ reaches an average AUROC of 84.2, improving over single-sample VC by $+12.9$ points (71.3$\rightarrow$84.2) and exceeding the best performance of either VC or SC even when scaled to $K{=}8$ (81.4 and 79.6, respectively), with gains far exceeding bootstrap variability. In STEM and humanities, SCVC estimated with two samples improves over single-sample VC by $+6.4$ AUROC (73.8$\rightarrow$80.2 and 68.5$\rightarrow$74.9) and surpasses the strongest single-signal estimates at $K{=}8$ by $+1.8$ and $+2.3$ AUROC, respectively. Further scaling of SCVC exhibits diminishing returns, but  remains feasible, particularly for Math: adding six more samples yields a +4.2 AUROC gain on Math and roughly +2 AUROC on STEM and Humanities.

\paragraph{Context.}
Hybrid uncertainty estimators that combine introspective confidence and cross-sample agreement have been explored previously in standard language models \cite{chen2023quantifying, xiong_can_2024, rivera2024combining, huang2024calibratinglongformgenerationslarge}. However, prior work did not systematically isolate the contribution of signal complementarity relative to increased sampling depth. This distinction was less relevant in standard language models, where per-sample token cost is lower and sampling is substantially more affordable than in RLMs.

Moreover, this behavior may have not been taking place under extended chain-of-thought reasoning: a single reasoning trace could already integrate over multiple hypotheses, potentially reducing the marginal value of cross-sample agreement \cite{podolak2025readmindreasoninghelps}. Our results show that this is not the case: introspection- and agreement-based signals remain strongly complementary even after extended deliberation.

\paragraph{Implication.}
As a result, the practical significance of the low-sample complementarity finding is amplified in reasoning language models, where each additional sample entails a long and costly chain-of-thought trace. Rather than allocating compute to deeper sampling of a single uncertainty signal, our results indicate that drawing a single additional sample and combining verbalized confidence with self-consistency yields the largest improvement per unit cost. This leads to a simple and robust recipe for uncertainty estimation in RLMs: avoid single-sample estimation, avoid pure self-consistency, and instead combine introspective confidence and cross-sample agreement using two samples.

\subsection{Finding III: Uncertainty signals scale fastest and combine most effectively in mathematics}
\label{sec:results:domain_dependance}

Uncertainty estimation exhibits strong domain dependence in both scaling speed and signal complementarity. Mathematics stands out as the domain where uncertainty signals improve most rapidly and combine most effectively. Moving from single-sample VC to the hybrid SCVC estimator at $K{=}2$ yields a large gain in mathematics (+12.9 AUROC), after which SCVC continues to improve from $K{=}2$ to $K{=}5$ (+2.7) and still gains from $K{=}5$ to $K{=}8$ (+1.5). In contrast, STEM and humanities show smaller initial gains (+6.4 AUROC) and much earlier saturation, with only $\approx$+1--2 AUROC improvement beyond $K{=}2$.

This pattern mirrors the behavior of individual signals: VC scales more strongly in mathematics, and SC reaches higher absolute quality and exhibits stronger late-stage scaling than in other domains. The alignment of faster scaling for VC, SC, and their combination indicates that uncertainty signals are both richer and more complementary in mathematical reasoning. While we do not isolate training causes, this behavior is consistent with reasoning language models being most effective and most extensively optimized in mathematics.

\subsection{Analysis}
\label{sec:analysis_ablations}

\paragraph{Combination of SC and VC is robust to weighting parameter $\lambda$}
\label{sec:analysis:weighting}

\begin{figure}[t]
    \centering
    \includegraphics[width=\linewidth]{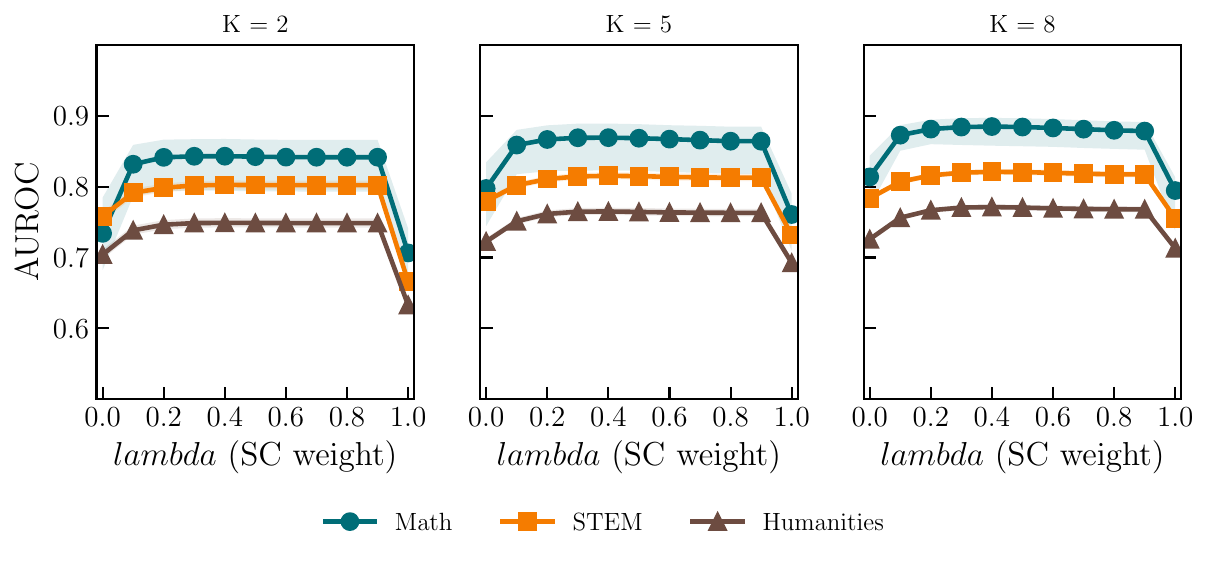}
    \caption{AUROC as a function of the SC weight $\lambda$ in the hybrid SC+VC signal, shown for $K{=}2$, $5$, and $8$. Results are averaged across models and tasks within each domain, with shaded regions indicating 95\% confidence intervals. Performance is stable across a wide range of $\lambda$, with degradation only at the extremes corresponding to pure VC or pure SC.}
    \label{fig:weight_sweep}
\end{figure}

So far, we combine VC and SC using an equal-weighted sum ($\lambda{=}0.5$). We test the sensitivity of this hybrid to the weighting parameter $\lambda \in [0,1]$, where $\lambda{=}0$ and $\lambda{=}1$ correspond to pure VC and pure SC, respectively.

Figure~\ref{fig:weight_sweep} shows AUROC as a function of $\lambda$ for $K \in \{2,5,8\}$, averaged across models and tasks within each domain. Across all domains and sampling budgets, SC+VC performance is largely invariant to $\lambda$ over a wide interior range: any non-degenerate combination ($0 < \lambda < 1$) yields nearly identical AUROC, with degradation only at the extremes where one signal is removed.

This robustness indicates that hybrid gains do not rely on precise weighting, but rather on the presence of both signals, making simple equal-weighted combination sufficient in practice.

\begin{figure}[t]
    \centering
    \includegraphics[width=0.7\linewidth]{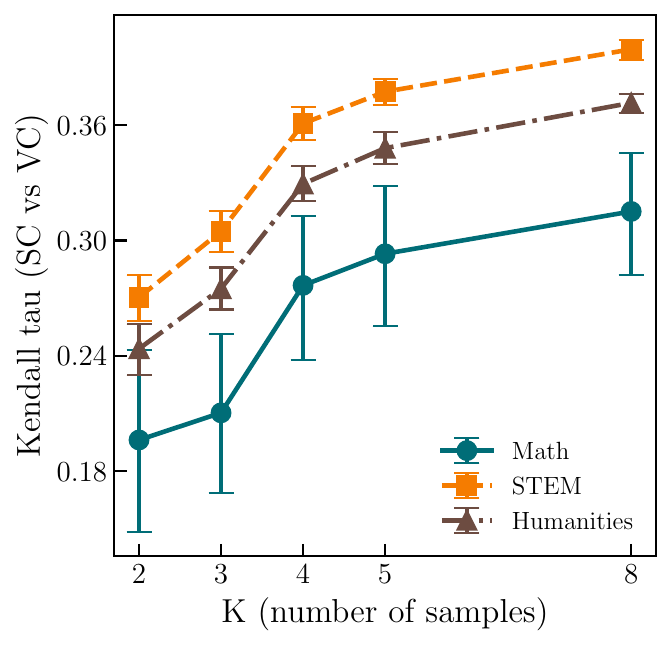}
    \caption{Kendall’s $\tau$ rank correlation between VC and SC as a function of the number of samples $K$ macro-averaged across reasoning models and task families. Correlation starts low and increases with sampling depth mirroring describing front-loaded gains of simple addition of the two signals and is consistently lower in mathematics than in STEM and humanities coinciding with RLVR training on math.}
    \label{fig:kendall_sc_vc}
\end{figure}

\paragraph{Complementarity can be described by correlation}
We analyze the relationship between verbalized confidence and self-consistency by measuring their rank correlation as a function of sampling budget $K$ and domain. Figure~\ref{fig:kendall_sc_vc} reports Kendall’s $\tau$ between VC and SC. Across all domains, the correlation increases monotonically with the number of samples, indicating that the two signals become progressively more aligned as additional samples are drawn. Moreover, the correlation is consistently higher in non-mathematical domains than in mathematics, particularly at small sampling budgets.

These regimes coincide with the behavior observed in Section~\ref{sec:results:complementarity}: hybrid gains are largest when correlation is lowest (early sampling and mathematics), and diminish as correlation increases (deeper sampling and non-RLVR-aligned domains). This shows that the benefit of hybrid uncertainty estimation does not arise from persistent signal independence. Instead, complementarity is strongest when VC and SC capture transiently distinct uncertainty information early in sampling, and weakens as both signals converge toward a shared notion of uncertainty with deeper sampling or outside RLVR-aligned domains.

\subsection{Revisiting verbalized confidence variants in reasoning language models}
\label{sec:vc_variants}

We study two families of verbalized confidence (VC) methods: \emph{elicitation} \citep{xiong_can_2024}, where the model reports confidence alongside its answer, and \emph{judge} approaches \citep{gu2025surveyllmasajudge}, where a separate pass reads the full reasoning trace and outputs a confidence score. Both estimate uncertainty via VC but differ in how the signal is extracted. We evaluate three instruction variants, each mapping responses to a 1–100 confidence scale (Appendix~\ref{app:prompts}), yielding six VC methods that incentivize different forms of introspection in reasoning language models (RLMs). While advanced variants showed limited success in short-trace LLMs, the longer chains of thought in RLMs may better support their assumptions, motivating a systematic comparison.

\textbf{\textit{Vanilla elicitation} (VaEl).}
The model provides an answer and a confidence score. This assumes direct introspective access to uncertainty and has been shown to work well in both LLMs and RLMs \citep{xiong_can_2024,yoon2025reasoning,zeng_reasoning_2025}.

\textbf{\textit{Verification elicitation} (VeEl).}
The model is prompted to check its reasoning before assigning confidence. While short LLM traces often lack structure, RLM scratchpads may better support self-verification \citep{miao2023selfcheckusingllmszeroshot}.

\textbf{\textit{Epistemic elicitation} (EpEl).}
The model is steered to reflect on its certainty during reasoning. This relies on extended reasoning budgets enabling simultaneous problem solving and self-assessment. Prior work found little benefit in LLMs \citep{tian_just_2023}, but longer RLM traces make this assumption more plausible.

\textbf{\textit{Vanilla judge} (VaJu).}
A second pass reads the full reasoning trace and outputs a confidence score. Sparse traces limited this signal in LLMs, while RLMs provide richer evidence \citep{xiong_can_2024}.

\textbf{\textit{Verification judge} (VeJu).}
The judge evaluates the validity and consistency of reasoning steps. This depends on explicit logical structure, which RLM traces expose more clearly than LLM outputs \citep{miao2023selfcheckusingllmszeroshot}.

\textbf{\textit{Epistemic-markers judge} (EpJu).}
The judge attends to hedging language and certainty cues. Such markers were unreliable in LLMs \citep{liu2025revisitingepistemicmarkersconfidence} but occur more frequently in extended RLM traces \citep{venhoff2025understanding}.

\paragraph{Result.}
This analysis uses the same three reasoning models as earlier sections; STEM includes GPQA and MMLU-Pro Physics and Health, and humanities include MMLU-Pro Psychology, Business, and Law.

Table~\ref{tab:domain_summary} shows that in mathematics, minimally guided methods perform best, with vanilla elicitation (VaEl) and verification judging (VeJu) achieving 73.2 and 73.8 AUROC, while epistemic variants underperform. In STEM, elicitation dominates, with VaEl highest at 77.2 AUROC. In humanities, differences narrow and judgment-based variants slightly outperform elicitation-based ones. Epistemic-marker judging is consistently weakest.

Judge differences arise despite identical reasoning traces, indicating sensitivity to confidence elicitation rather than information availability. Moreover, SCVC with two samples delivers substantially larger gains than judge-based methods while avoiding an extra reasoning-model pass, making judges an unfavorable cost–benefit tradeoff in RLMs.

\section{Conclusion}

We studied how uncertainty estimation scaled with parallel sampling in reasoning LMs using black-box methods. Both verbalized confidence and self-consistency improved with sampling, but self-consistency consistently lagged. Most gains arose from signal complementarity: combining both signals with just two samples outperformed deeper sampling of either alone. Effects were domain-dependent, with the strongest gains in mathematics, while advanced VC variants offered little or no benefit. These insights provide practical guidance for sampling-based uncertainty estimation in reasoning models, where extra samples are costly.
\begin{table}[t]
\centering
\begin{tabular}{lccc}
\toprule
Method & Math & STEM & Humanities \\
\midrule
VaEl & 73.24 & \textbf{77.15} & 70.04 \\
VeEl & 70.80 & 76.95 & 70.61 \\
EpEl & 69.44 & 76.11 & 71.43 \\
VaJu & 70.35 & 76.24 & \textbf{71.78} \\
VeJu & \textbf{73.82} & 76.37 & 71.49 \\
EpJu & 67.59 & 72.00 & 67.97 \\
\bottomrule
\end{tabular}
\caption{Domain-averaged AUROC for six uncertainty elicitation methods. Math includes AIME, GSM8K, and MMLU-Pro Math; STEM includes Physics, Health, and GPQA; Humanities includes remaining MMLU-Pro tasks.}
\label{tab:domain_summary}
\end{table}


\bibliography{custom,ConfidenceLRMs,main}

\appendix

\newpage
\section{Appendix: Breakdown of AUROC and Accuracy values per tasks and models}
\label{app:breakdown}


Tables~\ref{tab:full-auroc-math}--\ref{tab:full-auroc-humanities} report the AUROC values for verbalized confidence, self-consistency, and their combination across multiple domains. Each table presents results stratified by model, task, and number of samples. Specifically, Table~\ref{tab:full-auroc-math} focuses on mathematics tasks, Table~\ref{tab:full-auroc-stem} on STEM tasks, and Table~\ref{tab:full-auroc-humanities} on humanities tasks.

Tables~\ref{tab:full-acc-math}--\ref{tab:full-acc-humanities} present the corresponding results in terms of accuracy, following the same structure and domain breakdown.

\begin{table*}[h!]
\centering

\input{tables_readonly/auroc_task_tables_combined_math}

\caption{AUROC of verbalized confidence (VC), self-consistency (SC), and their combination (SCVC) across \textbf{mathematics} tasks at different sampling budgets $K$. Rows report per-model results, with the “Average (models)” row denoting the mean $\pm$ bootstrap standard deviation after averaging across models.}
\label{tab:full-auroc-math}
\end{table*}

\begin{table*}[h!]
\centering

\input{tables_readonly/auroc_task_tables_combined_ste}

\caption{AUROC of verbalized confidence (VC), self-consistency (SC), and their combination (SCVC) across \textbf{STEM} tasks at different sampling budgets $K$. Rows report per-model results, with the “Average (models)” row denoting the mean $\pm$ bootstrap standard deviation after averaging across models..}
\label{tab:full-auroc-stem}
\end{table*}

\begin{table*}[h!]
\centering

\input{tables_readonly/auroc_task_tables_combined_humanities}

\caption{AUROC of verbalized confidence (VC), self-consistency (SC), and their combination (SCVC) across \textbf{humanities} tasks at different sampling budgets $K$. Rows report per-model results, with the “Average (models)” row denoting the mean $\pm$ bootstrap standard deviation after averaging across models.}
\label{tab:full-auroc-humanities}
\end{table*}


\begin{table*}[h!]
\centering

\input{tables_readonly/acc_task_tables_combined_math}

\caption{Accuracy of verbalized confidence (VC), self-consistency (SC), and their combination (SCVC) across \textbf{mathematics} tasks at different sampling budgets $K$. Rows report per-model results, with the “Average (models)” row denoting the mean $\pm$ bootstrap standard deviation after averaging across models.}
\label{tab:full-acc-math}
\end{table*}

\begin{table*}[h!]
\centering

\input{tables_readonly/acc_task_tables_combined_ste}

\caption{Accuracy of verbalized confidence (VC), self-consistency (SC), and their combination (SCVC) across \textbf{STEM} tasks at different sampling budgets $K$. Rows report per-model results, with the “Average (models)” row denoting the mean $\pm$ bootstrap standard deviation after averaging across models.}
\label{tab:full-acc-stem}
\end{table*}

\begin{table*}[h!]
\centering

\input{tables_readonly/acc_task_tables_combined_humanities}

\caption{Accuracy of verbalized confidence (VC), self-consistency (SC), and their combination (SCVC) across \textbf{humanities} tasks at different sampling budgets $K$. Rows report per-model results, with the “Average (models)” row denoting the mean $\pm$ bootstrap standard deviation after averaging across models.}
\label{tab:full-acc-humanities}
\end{table*}

\clearpage
\section{Appendix: Detailed prompts}
\label{app:prompts}

\begin{figure}[H]
    \centering    
    \includegraphics[width=1\linewidth]{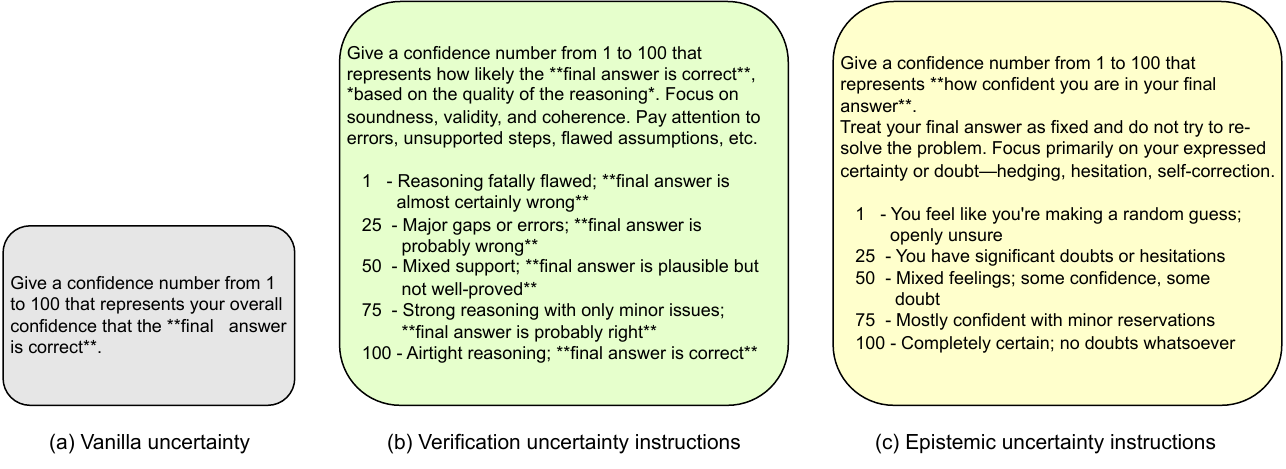}
    \caption{Overview of uncertainty instructions prompts defining VC methods. (a) Vanilla uncertainty instruction, (b) Verification uncertainty instruction and (c) epistemic uncertainty instruction. Each of the instructions is used both for elicitation and judge methods. For judge method, the epistemic uncertainty instructions are a bit different, as it needs to pay attention to the solver's reasoning trace, not its own.
    }

    \label{fig:prompts-overview}
\end{figure}

This Section describes the detailed prompts, and Figure~\ref{fig:prompts-overview} provides an overview of those used to obtain answers and elicit confidence.

Next are given exact prompt descriptions. Prompt~\ref{pr:elicit} is used for getting model-elicited uncertainties. Prompt~\ref{pr:nounc} is used to get the LRM thought trace without uncertainties, and after prompt~\ref{pr:judge} is used as a judge, giving us the Judge method. Prompts~\ref{pr:elicit}, \ref{pr:nounc} and \ref{pr:judge} are used for multiple-answer questions. Prompts~\ref{pr:elicit_math}, \ref{pr:nounc_math} and \ref{pr:judge_math} are like prompts~\ref{pr:elicit}, \ref{pr:nounc} and \ref{pr:judge}, but for the math dataset.

Inside the prompts~\ref{pr:elicit}, \ref{pr:nounc} and \ref{pr:judge} (same for \ref{pr:elicit_math}, \ref{pr:nounc_math} and \ref{pr:judge_math}), there are brackets which are used for inputting variables. \textit{Question}, \textit{choices}, and \textit{letter} correspond to the question in hand from the dataset, but \textit{uncertainty\_instructions} are subprompts which are based on the method type: vanilla (prompt~\ref{pr:unc_vanilla}), verification (prompt~\ref{pr:unc_verif}) and epistemic for the elicitation and the judge method (prompt~\ref{pr:unc_epi_el} and \ref{pr:unc_epi_judge}).

\begin{lstlisting}[style=prompt, 
caption={LRM prompt for multiple choice question confidence elicitation}, label={pr:elicit},
float=tbp,]
You are given a multiple choice question. 

**Solve the problem**, showing your reasoning step by step. After solving, provide your confidence in your answer.

{uncertainty_instructions}

{question}

{choices}

Your response must *end* with exactly two lines of the *exact* format below (no quotes) as the very final lines of your answer:
'ANSWER: $LETTER'  
'CONFIDENCE: $NUMBER'

- $LETTER must be one of the following options: {letters}

Think step by step before answering and show your reasoning first.
\end{lstlisting}

\begin{lstlisting}[style=prompt,
caption={LRM prompt for multiple choice question without uncertainty elicitation}, label={pr:nounc},
float=tbp]
You are given a multiple choice question. 

Solve the problem, showing your reasoning step by step.

{question}

{choices}

Your response must *end* with exactly one line of the *exact* format below (no quotes) as the very final line of your answer:
'ANSWER: $LETTER'

- $LETTER must be one of the following options: {letters}

Think step by step before answering and show your reasoning first.
\end{lstlisting}

\begin{lstlisting}[style=prompt,
caption={LRM prompt for multiple choice question for judging solver's reasoning process}, label={pr:judge},
float=tbp]
You are given a problem along with a solver's full reasoning process and the answer they arrived at. 

{uncertainty_instructions}

[BEGIN PROBLEM]
{question}
[END PROBLEM]

[BEGIN REASONING AND ANSWER]
{reasoning_and_answer}
[END REASONING AND ANSWER]

Think step by step. End your final answer with this exact format as the final line:
CONFIDENCE: [number between 1 and 100]
\end{lstlisting}

\begin{lstlisting}[style=prompt,
caption={Vanilla uncertainty prompt for uncertainty instructions bracket in the main prompt}, label={pr:unc_vanilla},
float=tbp]
Give a confidence number from 1 to 100 that represents your overall confidence that the **final answer is correct**.
\end{lstlisting}

\begin{lstlisting}[style=prompt,
caption={Verification uncertainty prompt for uncertainty instructions bracket in the main prompt}, label={pr:unc_verif},
float=tbp]
Give a confidence number from 1 to 100 that represents how likely the **final answer is correct**, *based on the quality of the reasoning*. Focus on soundness, validity, and coherence. Pay attention to errors, unsupported steps, flawed assumptions, etc.

    1   - Reasoning fatally flawed; **final answer is almost certainly wrong**  
    25  - Major gaps or errors; **final answer is probably wrong**  
    50  - Mixed support; **final answer is plausible but not well-proved**  
    75  - Strong reasoning with only minor issues; **final answer is probably right**  
    100 - Airtight reasoning; **final answer is correct**
\end{lstlisting}

\begin{lstlisting}[style=prompt,
caption={Epistemic uncertainty prompt for uncertainty instructions bracket in the main prompt for elicitation}, label={pr:unc_epi_el},
float=tbp]
Give a confidence number from 1 to 100 that represents **how confident you are in your final answer**.
Treat your final answer as fixed and do not try to re-solve the problem. Focus primarily on your expressed certainty or doubt-hedging, hesitation, self-correction.

    1   - You feel like you're making a random guess; openly unsure  
    25  - You have significant doubts or hesitations  
    50  - Mixed feelings; some confidence, some doubt  
    75  - Mostly confident with minor reservations  
    100 - Completely certain; no doubts whatsoever
\end{lstlisting}

\begin{lstlisting}[style=prompt,
caption={Epistemic uncertainty prompt for uncertainty instructions bracket in the main prompt for judge}, label={pr:unc_epi_judge},
float=tbp]
Give a confidence number from 1 to 100 that represents **how confident the solver is in their final answer**.
Treat their final answer as fixed and do not try to re-solve the problem. Focus primarily on their expressed certainty or doubt-hedging, hesitation, self-correction.

    1   - They feel like they're making a random guess; openly unsure  
    25  - They have significant doubts or hesitations  
    50  - Mixed feelings; some confidence, some doubt  
    75  - Mostly confident with minor reservations  
    100 - Completely certain; no doubts whatsoever
\end{lstlisting}

\begin{lstlisting}[style=prompt, 
caption={LRM prompt for math question confidence elicitation}, label={pr:elicit_math},
float=tbp]
You are given a math problem.

**Solve the problem**, showing your reasoning step by step. After solving, provide your confidence in your answer.

{uncertainty_instructions}

{prompt}

Your response must *end* with exactly two lines of the *exact* format below (no quotes) as the very final lines of your answer:
'ANSWER: $ANSWER'  
'CONFIDENCE: $NUMBER'

- Do not use LaTeX boxes like \boxed in the final lines; output plain text only.
- Think step by step before answering and show your reasoning first.
\end{lstlisting}

\begin{lstlisting}[style=prompt, 
caption={LRM prompt for math question without uncertainty elicitation}, label={pr:nounc_math},
float=tbp]
You are given a math problem.

Solve the problem, showing your reasoning step by step.

{prompt}

Your response must *end* with exactly one line of the *exact* format below (no quotes) as the very final line of your answer:
'ANSWER: $ANSWER'

- Do not use LaTeX boxes like \boxed in the final line; output plain text only.
- Think step by step before answering and show your reasoning first.
\end{lstlisting}

\begin{lstlisting}[style=prompt, 
caption={LRM prompt for math question for judging solver's reasoning process}, label={pr:judge_math},
float=tbp]
You are given a problem along with a solver's full reasoning process and the answer they arrived at. 

{uncertainty_instructions}

[BEGIN PROBLEM]
{question}
[END PROBLEM]

[BEGIN REASONING AND ANSWER]
{reasoning_and_answer}
[END REASONING AND ANSWER]

Think step by step. End your final answer with this exact format as the final line:
CONFIDENCE: [number between 1 and 100]

- Do not use LaTeX boxes like \boxed in the final line; output plain text only.
\end{lstlisting}

\end{document}

%% file: tables_readonly/task_families_examples.tex
{\scriptsize
\begin{tabular}{ll}
\toprule
Task & \# examples \\
\midrule
\multicolumn{2}{c}{\textbf{Mathematical tasks}} \\
Math & 1351 \\
GSM8K & 1319 \\
AIME 2024 \& 2025 & 60 \\
\multicolumn{2}{c}{\textbf{STEM tasks (excluding math)}} \\
Health & 818 \\
Biology & 717 \\
Chemistry & 1132 \\
Economics & 844 \\
Physics & 1299 \\
Computer Science & 410 \\
Engineering & 969 \\
GPQA Diamond & 198 \\
\multicolumn{2}{c}{\textbf{Humanities tasks}} \\
Psychology & 798 \\
Law & 1101 \\
Business & 789 \\
History & 381 \\
Philosophy & 499 \\
Other & 924 \\
\bottomrule
\end{tabular}
}

%% file: tables_readonly/auroc_task_tables_combined_summary_tasks.tex
{\scriptsize
\begin{tabular}{l>{\columncolor{green!10}}c>{\columncolor{green!10}}c>{\columncolor{green!10}}c>{\columncolor{green!10}}c>{\columncolor{blue!8}}c>{\columncolor{blue!8}}c>{\columncolor{blue!8}}c>{\columncolor{red!8}}c>{\columncolor{red!8}}c>{\columncolor{red!8}}c}
\toprule
 & \tiny VC (K=1) & \tiny VC (K=2) & \tiny VC (K=5) & \tiny VC (K=8) & \tiny SC (K=2) & \tiny SC (K=5) & \tiny SC (K=8) & \tiny SCVC (K=2) & \tiny SCVC (K=5) & \tiny SCVC (K=8) \\
\midrule
\multicolumn{11}{c}{Mathematical tasks} \\
Math & 78.7{\tiny $\tpm$1.0} & 81.1{\tiny $\tpm$1.0} & 84.5{\tiny $\tpm$0.6} & 84.9{\tiny $\tpm$0.5} & 66.1{\tiny $\tpm$1.2} & 74.1{\tiny $\tpm$1.1} & 77.6{\tiny $\tpm$0.9} & \textbf{85.5{\tiny $\tpm$0.8}} & 87.9{\tiny $\tpm$0.6} & 88.7{\tiny $\tpm$0.4} \\
GSM8K & 68.7{\tiny $\tpm$1.3} & 71.9{\tiny $\tpm$1.3} & 75.9{\tiny $\tpm$0.9} & 77.5{\tiny $\tpm$0.6} & 63.3{\tiny $\tpm$1.2} & 68.2{\tiny $\tpm$1.0} & 70.7{\tiny $\tpm$0.6} & \textbf{77.7{\tiny $\tpm$1.2}} & 80.8{\tiny $\tpm$0.8} & 82.4{\tiny $\tpm$0.5} \\
AIME 2024 \& 2025 & 66.4{\tiny $\tpm$7.6} & 67.2{\tiny $\tpm$8.2} & 78.8{\tiny $\tpm$6.8} & 81.7{\tiny $\tpm$5.5} & 82.6{\tiny $\tpm$6.1} & 85.9{\tiny $\tpm$7.9} & 90.0{\tiny $\tpm$4.9} & \textbf{89.5{\tiny $\tpm$5.5}} & 91.9{\tiny $\tpm$5.1} & 94.1{\tiny $\tpm$2.7} \\
Average (tasks) & \cellcolor{green!18}71.3{\tiny $\tpm$2.6} & \cellcolor{green!18}73.4{\tiny $\tpm$2.8} & \cellcolor{green!18}79.7{\tiny $\tpm$2.3} & \cellcolor{green!18}81.4{\tiny $\tpm$1.9} & \cellcolor{blue!14}70.6{\tiny $\tpm$2.1} & \cellcolor{blue!14}76.1{\tiny $\tpm$2.6} & \cellcolor{blue!14}79.4{\tiny $\tpm$1.6} & \cellcolor{red!14}\textbf{84.2{\tiny $\tpm$1.9}} & \cellcolor{red!14}86.8{\tiny $\tpm$1.7} & \cellcolor{red!14}88.4{\tiny $\tpm$0.9}\\ 
\multicolumn{11}{c}{STEM tasks (excluding math)} \\
Health & 69.9{\tiny $\tpm$0.9} & 71.9{\tiny $\tpm$0.8} & 73.8{\tiny $\tpm$0.5} & 74.3{\tiny $\tpm$0.3} & 63.4{\tiny $\tpm$0.8} & 69.5{\tiny $\tpm$0.7} & 71.8{\tiny $\tpm$0.5} & \textbf{76.2{\tiny $\tpm$0.8}} & 77.4{\tiny $\tpm$0.5} & 77.9{\tiny $\tpm$0.4} \\
Biology & 72.4{\tiny $\tpm$1.1} & 74.5{\tiny $\tpm$1.1} & 76.6{\tiny $\tpm$0.6} & 77.1{\tiny $\tpm$0.4} & 64.8{\tiny $\tpm$1.2} & 71.7{\tiny $\tpm$1.0} & 74.5{\tiny $\tpm$0.6} & \textbf{79.4{\tiny $\tpm$1.0}} & 81.0{\tiny $\tpm$0.7} & 81.7{\tiny $\tpm$0.4} \\
Chemistry & 78.0{\tiny $\tpm$0.8} & 80.2{\tiny $\tpm$0.8} & 82.6{\tiny $\tpm$0.5} & 83.0{\tiny $\tpm$0.4} & 67.4{\tiny $\tpm$1.0} & 75.3{\tiny $\tpm$0.8} & 78.1{\tiny $\tpm$0.6} & \textbf{83.7{\tiny $\tpm$0.7}} & 85.6{\tiny $\tpm$0.5} & 86.2{\tiny $\tpm$0.3} \\
Economics & 69.7{\tiny $\tpm$0.9} & 71.8{\tiny $\tpm$0.9} & 73.7{\tiny $\tpm$0.5} & 74.1{\tiny $\tpm$0.4} & 63.2{\tiny $\tpm$1.0} & 70.0{\tiny $\tpm$0.8} & 72.6{\tiny $\tpm$0.5} & \textbf{75.9{\tiny $\tpm$0.9}} & 77.9{\tiny $\tpm$0.6} & 78.7{\tiny $\tpm$0.4} \\
Physics & 74.6{\tiny $\tpm$0.8} & 76.1{\tiny $\tpm$0.8} & 78.3{\tiny $\tpm$0.6} & 78.6{\tiny $\tpm$0.5} & 70.6{\tiny $\tpm$0.9} & 77.0{\tiny $\tpm$0.7} & 79.4{\tiny $\tpm$0.6} & \textbf{82.3{\tiny $\tpm$0.7}} & 83.3{\tiny $\tpm$0.5} & 83.8{\tiny $\tpm$0.4} \\
Computer Science & 72.0{\tiny $\tpm$1.3} & 74.2{\tiny $\tpm$1.2} & 75.7{\tiny $\tpm$0.8} & 76.0{\tiny $\tpm$0.5} & 62.3{\tiny $\tpm$1.4} & 66.7{\tiny $\tpm$1.2} & 68.8{\tiny $\tpm$0.8} & \textbf{77.4{\tiny $\tpm$1.2}} & 78.1{\tiny $\tpm$0.9} & 78.7{\tiny $\tpm$0.6} \\
Engineering & 79.2{\tiny $\tpm$0.6} & 80.9{\tiny $\tpm$0.6} & 82.5{\tiny $\tpm$0.4} & 83.0{\tiny $\tpm$0.3} & 74.6{\tiny $\tpm$0.8} & 81.9{\tiny $\tpm$0.5} & 83.5{\tiny $\tpm$0.4} & \textbf{86.5{\tiny $\tpm$0.5}} & 87.0{\tiny $\tpm$0.4} & 87.3{\tiny $\tpm$0.3} \\
GPQA Diamond & 74.6{\tiny $\tpm$1.6} & 76.8{\tiny $\tpm$1.6} & 80.2{\tiny $\tpm$1.3} & 80.8{\tiny $\tpm$1.2} & 66.3{\tiny $\tpm$1.8} & 73.3{\tiny $\tpm$1.7} & 75.4{\tiny $\tpm$1.5} & \textbf{80.3{\tiny $\tpm$1.5}} & 81.6{\tiny $\tpm$1.3} & 82.1{\tiny $\tpm$1.2} \\
Average (tasks) & \cellcolor{green!18}73.8{\tiny $\tpm$0.4} & \cellcolor{green!18}75.8{\tiny $\tpm$0.4} & \cellcolor{green!18}77.9{\tiny $\tpm$0.3} & \cellcolor{green!18}78.3{\tiny $\tpm$0.2} & \cellcolor{blue!14}66.6{\tiny $\tpm$0.4} & \cellcolor{blue!14}73.2{\tiny $\tpm$0.3} & \cellcolor{blue!14}75.5{\tiny $\tpm$0.3} & \cellcolor{red!14}\textbf{80.2{\tiny $\tpm$0.3}} & \cellcolor{red!14}81.5{\tiny $\tpm$0.3} & \cellcolor{red!14}82.0{\tiny $\tpm$0.2}\\ 
\multicolumn{11}{c}{Humanities tasks} \\
Psychology & 68.1{\tiny $\tpm$1.0} & 70.4{\tiny $\tpm$0.9} & 72.9{\tiny $\tpm$0.7} & 73.7{\tiny $\tpm$0.6} & 63.3{\tiny $\tpm$0.8} & 70.0{\tiny $\tpm$0.8} & 72.5{\tiny $\tpm$0.8} & \textbf{75.5{\tiny $\tpm$0.8}} & 77.7{\tiny $\tpm$0.7} & 78.6{\tiny $\tpm$0.6} \\
Law & 57.4{\tiny $\tpm$0.8} & 58.5{\tiny $\tpm$0.7} & 59.6{\tiny $\tpm$0.5} & 59.4{\tiny $\tpm$0.4} & 61.4{\tiny $\tpm$0.7} & 65.5{\tiny $\tpm$0.6} & 66.8{\tiny $\tpm$0.5} & \textbf{65.4{\tiny $\tpm$0.8}} & 67.4{\tiny $\tpm$0.6} & 68.1{\tiny $\tpm$0.5} \\
Business & 79.2{\tiny $\tpm$0.8} & 81.2{\tiny $\tpm$0.8} & 83.0{\tiny $\tpm$0.5} & 83.2{\tiny $\tpm$0.3} & 66.3{\tiny $\tpm$1.1} & 73.7{\tiny $\tpm$0.8} & 76.3{\tiny $\tpm$0.5} & \textbf{84.3{\tiny $\tpm$0.7}} & 85.6{\tiny $\tpm$0.5} & 86.1{\tiny $\tpm$0.3} \\
History & 63.6{\tiny $\tpm$1.1} & 65.4{\tiny $\tpm$1.1} & 67.3{\tiny $\tpm$0.7} & 67.5{\tiny $\tpm$0.5} & 60.0{\tiny $\tpm$1.0} & 64.8{\tiny $\tpm$0.9} & 66.6{\tiny $\tpm$0.6} & \textbf{69.3{\tiny $\tpm$1.1}} & 70.9{\tiny $\tpm$0.9} & 71.6{\tiny $\tpm$0.6} \\
Philosophy & 71.7{\tiny $\tpm$1.0} & 73.7{\tiny $\tpm$0.9} & 75.5{\tiny $\tpm$0.6} & 75.9{\tiny $\tpm$0.4} & 63.5{\tiny $\tpm$1.0} & 69.9{\tiny $\tpm$0.8} & 71.7{\tiny $\tpm$0.6} & \textbf{76.9{\tiny $\tpm$0.9}} & 77.8{\tiny $\tpm$0.7} & 78.2{\tiny $\tpm$0.5} \\
Other & 71.3{\tiny $\tpm$0.7} & 73.2{\tiny $\tpm$0.7} & 75.3{\tiny $\tpm$0.4} & 75.7{\tiny $\tpm$0.3} & 65.2{\tiny $\tpm$0.8} & 71.7{\tiny $\tpm$0.6} & 73.9{\tiny $\tpm$0.4} & \textbf{77.8{\tiny $\tpm$0.7}} & 79.1{\tiny $\tpm$0.5} & 79.7{\tiny $\tpm$0.4} \\
Average (tasks) & \cellcolor{green!18}68.5{\tiny $\tpm$0.4} & \cellcolor{green!18}70.4{\tiny $\tpm$0.4} & \cellcolor{green!18}72.3{\tiny $\tpm$0.2} & \cellcolor{green!18}72.6{\tiny $\tpm$0.2} & \cellcolor{blue!14}63.3{\tiny $\tpm$0.4} & \cellcolor{blue!14}69.3{\tiny $\tpm$0.3} & \cellcolor{blue!14}71.3{\tiny $\tpm$0.2} & \cellcolor{red!14}\textbf{74.9{\tiny $\tpm$0.3}} & \cellcolor{red!14}76.4{\tiny $\tpm$0.3} & \cellcolor{red!14}77.0{\tiny $\tpm$0.2}\\ 
\bottomrule
\end{tabular}
}

%% file: tables_readonly/auroc_task_tables_combined_math.tex
{\scriptsize
\begin{tabular}{l>{\columncolor{green!10}}c>{\columncolor{green!10}}c>{\columncolor{green!10}}c>{\columncolor{green!10}}c>{\columncolor{blue!8}}c>{\columncolor{blue!8}}c>{\columncolor{blue!8}}c>{\columncolor{red!8}}c>{\columncolor{red!8}}c>{\columncolor{red!8}}c}
\toprule
 & \tiny VC (K=1) & \tiny VC (K=2) & \tiny VC (K=5) & \tiny VC (K=8) & \tiny SC (K=2) & \tiny SC (K=5) & \tiny SC (K=8) & \tiny SCVC (K=2) & \tiny SCVC (K=5) & \tiny SCVC (K=8) \\
\midrule
\multicolumn{11}{c}{Math} \\
gpt-oss-20b & 80.9{\tiny $\tpm$2.1} & 83.2{\tiny $\tpm$1.9} & 87.9{\tiny $\tpm$1.0} & 88.6{\tiny $\tpm$0.7} & 63.6{\tiny $\tpm$2.1} & 71.5{\tiny $\tpm$2.0} & 75.2{\tiny $\tpm$1.5} & \textbf{87.1{\tiny $\tpm$1.5}} & 89.4{\tiny $\tpm$0.9} & 90.0{\tiny $\tpm$0.8} \\
Qwen3-30B-A3B & 76.3{\tiny $\tpm$1.7} & 79.4{\tiny $\tpm$1.5} & 81.8{\tiny $\tpm$1.0} & 81.6{\tiny $\tpm$0.7} & 62.8{\tiny $\tpm$2.1} & 70.6{\tiny $\tpm$1.9} & 74.1{\tiny $\tpm$1.7} & \textbf{82.2{\tiny $\tpm$1.5}} & 85.2{\tiny $\tpm$1.1} & 86.5{\tiny $\tpm$0.8} \\
DeepSeek-R1-8B & 78.9{\tiny $\tpm$1.8} & 80.7{\tiny $\tpm$1.7} & 83.8{\tiny $\tpm$1.2} & 84.5{\tiny $\tpm$0.9} & 71.8{\tiny $\tpm$2.0} & 80.1{\tiny $\tpm$1.6} & 83.4{\tiny $\tpm$1.4} & \textbf{87.2{\tiny $\tpm$1.3}} & 89.0{\tiny $\tpm$1.0} & 89.8{\tiny $\tpm$0.7} \\
Average (models) & \cellcolor{green!18}78.7{\tiny $\tpm$1.0} & \cellcolor{green!18}81.1{\tiny $\tpm$1.0} & \cellcolor{green!18}84.5{\tiny $\tpm$0.6} & \cellcolor{green!18}84.9{\tiny $\tpm$0.5} & \cellcolor{blue!14}66.1{\tiny $\tpm$1.2} & \cellcolor{blue!14}74.1{\tiny $\tpm$1.1} & \cellcolor{blue!14}77.6{\tiny $\tpm$0.9} & \cellcolor{red!14}\textbf{85.5{\tiny $\tpm$0.8}} & \cellcolor{red!14}87.9{\tiny $\tpm$0.6} & \cellcolor{red!14}88.7{\tiny $\tpm$0.4}\\ 
\multicolumn{11}{c}{GSM8K} \\
\bottomrule
\end{tabular}
}

%% file: tables_readonly/auroc_task_tables_combined_ste.tex
{\scriptsize
\begin{tabular}{l>{\columncolor{green!10}}c>{\columncolor{green!10}}c>{\columncolor{green!10}}c>{\columncolor{green!10}}c>{\columncolor{blue!8}}c>{\columncolor{blue!8}}c>{\columncolor{blue!8}}c>{\columncolor{red!8}}c>{\columncolor{red!8}}c>{\columncolor{red!8}}c}
\toprule
 & \tiny VC (K=1) & \tiny VC (K=2) & \tiny VC (K=5) & \tiny VC (K=8) & \tiny SC (K=2) & \tiny SC (K=5) & \tiny SC (K=8) & \tiny SCVC (K=2) & \tiny SCVC (K=5) & \tiny SCVC (K=8) \\
\midrule
\multicolumn{11}{c}{Health} \\
gpt-oss-20b & 74.3{\tiny $\tpm$1.4} & 76.4{\tiny $\tpm$1.3} & 78.6{\tiny $\tpm$0.8} & 79.1{\tiny $\tpm$0.6} & 64.4{\tiny $\tpm$1.5} & 71.2{\tiny $\tpm$1.3} & 73.9{\tiny $\tpm$0.9} & \textbf{79.8{\tiny $\tpm$1.2}} & 80.3{\tiny $\tpm$0.9} & 80.6{\tiny $\tpm$0.7} \\
Qwen3-30B-A3B & 67.4{\tiny $\tpm$1.6} & 69.7{\tiny $\tpm$1.4} & 72.8{\tiny $\tpm$0.9} & 73.3{\tiny $\tpm$0.6} & 62.1{\tiny $\tpm$1.5} & 68.7{\tiny $\tpm$1.1} & 71.0{\tiny $\tpm$0.8} & \textbf{74.3{\tiny $\tpm$1.4}} & 76.7{\tiny $\tpm$0.9} & 77.4{\tiny $\tpm$0.7} \\
DeepSeek-R1-8B & 67.9{\tiny $\tpm$1.3} & 69.7{\tiny $\tpm$1.2} & 70.1{\tiny $\tpm$0.8} & 70.4{\tiny $\tpm$0.6} & 63.6{\tiny $\tpm$1.3} & 68.5{\tiny $\tpm$1.2} & 70.4{\tiny $\tpm$0.9} & \textbf{74.6{\tiny $\tpm$1.2}} & 75.2{\tiny $\tpm$0.9} & 75.6{\tiny $\tpm$0.7} \\
Average (models) & \cellcolor{green!18}69.9{\tiny $\tpm$0.9} & \cellcolor{green!18}71.9{\tiny $\tpm$0.8} & \cellcolor{green!18}73.8{\tiny $\tpm$0.5} & \cellcolor{green!18}74.3{\tiny $\tpm$0.3} & \cellcolor{blue!14}63.4{\tiny $\tpm$0.8} & \cellcolor{blue!14}69.5{\tiny $\tpm$0.7} & \cellcolor{blue!14}71.8{\tiny $\tpm$0.5} & \cellcolor{red!14}\textbf{76.2{\tiny $\tpm$0.8}} & \cellcolor{red!14}77.4{\tiny $\tpm$0.5} & \cellcolor{red!14}77.9{\tiny $\tpm$0.4}\\ 
\multicolumn{11}{c}{Biology} \\
\bottomrule
\end{tabular}
}

%% file: tables_readonly/auroc_task_tables_combined_humanities.tex
{\scriptsize
\begin{tabular}{l>{\columncolor{green!10}}c>{\columncolor{green!10}}c>{\columncolor{green!10}}c>{\columncolor{green!10}}c>{\columncolor{blue!8}}c>{\columncolor{blue!8}}c>{\columncolor{blue!8}}c>{\columncolor{red!8}}c>{\columncolor{red!8}}c>{\columncolor{red!8}}c}
\toprule
 & \tiny VC (K=1) & \tiny VC (K=2) & \tiny VC (K=5) & \tiny VC (K=8) & \tiny SC (K=2) & \tiny SC (K=5) & \tiny SC (K=8) & \tiny SCVC (K=2) & \tiny SCVC (K=5) & \tiny SCVC (K=8) \\
\midrule
\multicolumn{11}{c}{Psychology} \\
gpt-oss-20b & 70.0{\tiny $\tpm$1.7} & 72.3{\tiny $\tpm$1.7} & 75.9{\tiny $\tpm$1.2} & 77.0{\tiny $\tpm$1.1} & 66.0{\tiny $\tpm$1.4} & 73.2{\tiny $\tpm$1.5} & 75.8{\tiny $\tpm$1.3} & \textbf{78.2{\tiny $\tpm$1.4}} & 80.0{\tiny $\tpm$1.1} & 80.8{\tiny $\tpm$1.0} \\
Qwen3-30B-A3B & 65.2{\tiny $\tpm$1.8} & 68.1{\tiny $\tpm$1.6} & 71.3{\tiny $\tpm$1.3} & 72.1{\tiny $\tpm$1.1} & 60.6{\tiny $\tpm$1.5} & 67.3{\tiny $\tpm$1.5} & 69.8{\tiny $\tpm$1.3} & \textbf{72.9{\tiny $\tpm$1.6}} & 76.7{\tiny $\tpm$1.3} & 77.9{\tiny $\tpm$1.1} \\
DeepSeek-R1-8B & 68.9{\tiny $\tpm$1.5} & 70.8{\tiny $\tpm$1.4} & 71.6{\tiny $\tpm$1.0} & 71.9{\tiny $\tpm$0.9} & 63.2{\tiny $\tpm$1.5} & 69.5{\tiny $\tpm$1.5} & 71.9{\tiny $\tpm$1.4} & \textbf{75.3{\tiny $\tpm$1.3}} & 76.5{\tiny $\tpm$1.1} & 77.1{\tiny $\tpm$1.0} \\
Average (models) & \cellcolor{green!18}68.1{\tiny $\tpm$1.0} & \cellcolor{green!18}70.4{\tiny $\tpm$0.9} & \cellcolor{green!18}72.9{\tiny $\tpm$0.7} & \cellcolor{green!18}73.7{\tiny $\tpm$0.6} & \cellcolor{blue!14}63.3{\tiny $\tpm$0.8} & \cellcolor{blue!14}70.0{\tiny $\tpm$0.8} & \cellcolor{blue!14}72.5{\tiny $\tpm$0.8} & \cellcolor{red!14}\textbf{75.5{\tiny $\tpm$0.8}} & \cellcolor{red!14}77.7{\tiny $\tpm$0.7} & \cellcolor{red!14}78.6{\tiny $\tpm$0.6}\\ 
\multicolumn{11}{c}{Law} \\
\bottomrule
\end{tabular}
}

%% file: tables_readonly/acc_task_tables_combined_math.tex
{\scriptsize
\begin{tabular}{l>{\columncolor{green!10}}c>{\columncolor{green!10}}c>{\columncolor{green!10}}c>{\columncolor{green!10}}c>{\columncolor{blue!8}}c>{\columncolor{blue!8}}c>{\columncolor{blue!8}}c>{\columncolor{red!8}}c>{\columncolor{red!8}}c>{\columncolor{red!8}}c}
\toprule
 & \tiny VC (K=1) & \tiny VC (K=2) & \tiny VC (K=5) & \tiny VC (K=8) & \tiny SC (K=2) & \tiny SC (K=5) & \tiny SC (K=8) & \tiny SCVC (K=2) & \tiny SCVC (K=5) & \tiny SCVC (K=8) \\
\midrule
\multicolumn{11}{c}{Math} \\
gpt-oss-20b & 94.1{\tiny $\tpm$0.3} & 94.0{\tiny $\tpm$0.3} & 94.5{\tiny $\tpm$0.2} & 94.5{\tiny $\tpm$0.1} & 94.0{\tiny $\tpm$0.3} & 94.5{\tiny $\tpm$0.2} & 94.5{\tiny $\tpm$0.1} & \textbf{94.0{\tiny $\tpm$0.3}} & 94.5{\tiny $\tpm$0.2} & 94.5{\tiny $\tpm$0.1} \\
Qwen3-30B-A3B & 94.5{\tiny $\tpm$0.2} & 94.5{\tiny $\tpm$0.2} & 94.7{\tiny $\tpm$0.2} & 94.7{\tiny $\tpm$0.1} & 94.5{\tiny $\tpm$0.2} & 94.7{\tiny $\tpm$0.2} & 94.7{\tiny $\tpm$0.1} & \textbf{94.5{\tiny $\tpm$0.2}} & 94.7{\tiny $\tpm$0.2} & 94.7{\tiny $\tpm$0.1} \\
DeepSeek-R1-8B & 91.8{\tiny $\tpm$0.4} & 91.8{\tiny $\tpm$0.4} & 92.8{\tiny $\tpm$0.2} & 92.9{\tiny $\tpm$0.2} & 91.8{\tiny $\tpm$0.4} & 92.8{\tiny $\tpm$0.2} & 92.9{\tiny $\tpm$0.2} & \textbf{91.8{\tiny $\tpm$0.4}} & 92.8{\tiny $\tpm$0.2} & 92.9{\tiny $\tpm$0.2} \\
Average (models) & \cellcolor{green!18}93.5{\tiny $\tpm$0.2} & \cellcolor{green!18}93.4{\tiny $\tpm$0.2} & \cellcolor{green!18}94.0{\tiny $\tpm$0.1} & \cellcolor{green!18}94.0{\tiny $\tpm$0.1} & \cellcolor{blue!14}93.4{\tiny $\tpm$0.2} & \cellcolor{blue!14}94.0{\tiny $\tpm$0.1} & \cellcolor{blue!14}94.0{\tiny $\tpm$0.1} & \cellcolor{red!14}\textbf{93.4{\tiny $\tpm$0.2}} & \cellcolor{red!14}94.0{\tiny $\tpm$0.1} & \cellcolor{red!14}94.0{\tiny $\tpm$0.1}\\ 
\multicolumn{11}{c}{GSM8K} \\
\bottomrule
\end{tabular}
}

%% file: tables_readonly/acc_task_tables_combined_ste.tex
{\scriptsize
\begin{tabular}{l>{\columncolor{green!10}}c>{\columncolor{green!10}}c>{\columncolor{green!10}}c>{\columncolor{green!10}}c>{\columncolor{blue!8}}c>{\columncolor{blue!8}}c>{\columncolor{blue!8}}c>{\columncolor{red!8}}c>{\columncolor{red!8}}c>{\columncolor{red!8}}c}
\toprule
 & \tiny VC (K=1) & \tiny VC (K=2) & \tiny VC (K=5) & \tiny VC (K=8) & \tiny SC (K=2) & \tiny SC (K=5) & \tiny SC (K=8) & \tiny SCVC (K=2) & \tiny SCVC (K=5) & \tiny SCVC (K=8) \\
\midrule
\multicolumn{11}{c}{Health} \\
gpt-oss-20b & 74.2{\tiny $\tpm$0.8} & 74.2{\tiny $\tpm$0.8} & 75.6{\tiny $\tpm$0.5} & 75.8{\tiny $\tpm$0.4} & 74.2{\tiny $\tpm$0.8} & 75.6{\tiny $\tpm$0.5} & 75.8{\tiny $\tpm$0.4} & \textbf{74.2{\tiny $\tpm$0.8}} & 75.6{\tiny $\tpm$0.5} & 75.8{\tiny $\tpm$0.4} \\
Qwen3-30B-A3B & 75.4{\tiny $\tpm$0.6} & 75.4{\tiny $\tpm$0.6} & 76.4{\tiny $\tpm$0.4} & 76.5{\tiny $\tpm$0.3} & 75.4{\tiny $\tpm$0.6} & 76.4{\tiny $\tpm$0.4} & 76.5{\tiny $\tpm$0.3} & \textbf{75.4{\tiny $\tpm$0.6}} & 76.4{\tiny $\tpm$0.4} & 76.5{\tiny $\tpm$0.3} \\
DeepSeek-R1-8B & 70.0{\tiny $\tpm$0.8} & 70.0{\tiny $\tpm$0.7} & 72.0{\tiny $\tpm$0.5} & 72.2{\tiny $\tpm$0.4} & 70.0{\tiny $\tpm$0.7} & 72.0{\tiny $\tpm$0.5} & 72.2{\tiny $\tpm$0.4} & \textbf{70.0{\tiny $\tpm$0.7}} & 72.0{\tiny $\tpm$0.5} & 72.2{\tiny $\tpm$0.4} \\
Average (models) & \cellcolor{green!18}73.2{\tiny $\tpm$0.4} & \cellcolor{green!18}73.2{\tiny $\tpm$0.4} & \cellcolor{green!18}74.7{\tiny $\tpm$0.3} & \cellcolor{green!18}74.8{\tiny $\tpm$0.2} & \cellcolor{blue!14}73.2{\tiny $\tpm$0.4} & \cellcolor{blue!14}74.7{\tiny $\tpm$0.3} & \cellcolor{blue!14}74.8{\tiny $\tpm$0.2} & \cellcolor{red!14}\textbf{73.2{\tiny $\tpm$0.4}} & \cellcolor{red!14}74.7{\tiny $\tpm$0.3} & \cellcolor{red!14}74.8{\tiny $\tpm$0.2}\\ 
\multicolumn{11}{c}{Biology} \\
\bottomrule
\end{tabular}
}

%% file: tables_readonly/acc_task_tables_combined_humanities.tex
{\scriptsize
\begin{tabular}{l>{\columncolor{green!10}}c>{\columncolor{green!10}}c>{\columncolor{green!10}}c>{\columncolor{green!10}}c>{\columncolor{blue!8}}c>{\columncolor{blue!8}}c>{\columncolor{blue!8}}c>{\columncolor{red!8}}c>{\columncolor{red!8}}c>{\columncolor{red!8}}c}
\toprule
 & \tiny VC (K=1) & \tiny VC (K=2) & \tiny VC (K=5) & \tiny VC (K=8) & \tiny SC (K=2) & \tiny SC (K=5) & \tiny SC (K=8) & \tiny SCVC (K=2) & \tiny SCVC (K=5) & \tiny SCVC (K=8) \\
\midrule
\multicolumn{11}{c}{Psychology} \\
gpt-oss-20b & 71.5{\tiny $\tpm$0.8} & 71.5{\tiny $\tpm$0.8} & 73.2{\tiny $\tpm$0.7} & 73.4{\tiny $\tpm$0.6} & 71.5{\tiny $\tpm$0.8} & 73.2{\tiny $\tpm$0.7} & 73.4{\tiny $\tpm$0.6} & \textbf{71.5{\tiny $\tpm$0.8}} & 73.2{\tiny $\tpm$0.7} & 73.4{\tiny $\tpm$0.6} \\
Qwen3-30B-A3B & 76.9{\tiny $\tpm$0.6} & 76.8{\tiny $\tpm$0.6} & 77.5{\tiny $\tpm$0.5} & 77.5{\tiny $\tpm$0.4} & 76.8{\tiny $\tpm$0.6} & 77.5{\tiny $\tpm$0.5} & 77.5{\tiny $\tpm$0.4} & \textbf{76.8{\tiny $\tpm$0.6}} & 77.5{\tiny $\tpm$0.5} & 77.5{\tiny $\tpm$0.4} \\
DeepSeek-R1-8B & 73.3{\tiny $\tpm$0.8} & 73.3{\tiny $\tpm$0.8} & 74.7{\tiny $\tpm$0.6} & 75.0{\tiny $\tpm$0.5} & 73.3{\tiny $\tpm$0.8} & 74.7{\tiny $\tpm$0.6} & 75.0{\tiny $\tpm$0.5} & \textbf{73.3{\tiny $\tpm$0.8}} & 74.7{\tiny $\tpm$0.6} & 75.0{\tiny $\tpm$0.5} \\
Average (models) & \cellcolor{green!18}73.9{\tiny $\tpm$0.4} & \cellcolor{green!18}73.9{\tiny $\tpm$0.4} & \cellcolor{green!18}75.1{\tiny $\tpm$0.3} & \cellcolor{green!18}75.3{\tiny $\tpm$0.3} & \cellcolor{blue!14}73.9{\tiny $\tpm$0.4} & \cellcolor{blue!14}75.1{\tiny $\tpm$0.3} & \cellcolor{blue!14}75.3{\tiny $\tpm$0.3} & \cellcolor{red!14}\textbf{73.9{\tiny $\tpm$0.4}} & \cellcolor{red!14}75.1{\tiny $\tpm$0.3} & \cellcolor{red!14}75.3{\tiny $\tpm$0.3}\\ 
\multicolumn{11}{c}{Law} \\
\bottomrule
\end{tabular}
}